\documentclass[runningheads]{llncs}
\usepackage{graphicx}
\usepackage{multirow}
\usepackage{multicol}
\usepackage{booktabs}
\usepackage{subfig}

\begin{document}
\title{Deep Class-specific Affinity-Guided Convolutional Network for Multimodal Unpaired Image Segmentation}
%
\titlerunning{Deep Class-specific Affinity-Guided Convolutional Network for Multimodal}
%
\author{Jingkun Chen\inst{1} \and
   Wenqi Li\inst{2} \and
   Hongwei Li \inst{3} \and
   Jianguo Zhang\inst{1} }

 \authorrunning{J. Chen et al.}
 \institute{Department of Computer Science and Engineering, \\ Southern University of Science and Technology, Shenzhen, China
\email{zhangjg@sustech.edu.cn}\and
NVIDIA \and
Technical University of Munich, Germany}
\maketitle              
\begin{abstract}

Multi-modal medical image segmentation plays an essential role in clinical diagnosis. It remains challenging as the input modalities are often not well-aligned spatially. Existing learning-based methods mainly consider sharing trainable layers across modalities and minimizing visual feature discrepancies.
While the problem is often formulated as joint supervised feature learning, multiple-scale features and class-specific representation have not yet been explored. In this paper, we propose an affinity-guided fully convolutional network for multimodal image segmentation. To learn effective representations, we design class-specific affinity matrices to encode the knowledge of hierarchical feature reasoning, together with the shared convolutional layers to ensure the cross-modality generalization. Our affinity matrix does not depend on spatial alignments of the visual features and thus allows us to train with unpaired, multimodal inputs. We extensively evaluated our method on two public multimodal benchmark datasets and outperform state-of-the-art methods.

\keywords{Segmentation \and Class-specific Affinity \and Feature Transfer.}
\end{abstract}

\section{Introduction}

Medical image segmentation is a key step in clinical diagnosis and
treatment. Fully convolutional networks
\cite{ronneberger2015unet,badrinarayanan2017segnet,long2015fcn} have been established as powerful tools for the
segmentation tasks.
Benefiting from the learning capability of these models, researchers start to
address more challenging and critical problems such as learning from multiple
imaging modalities. This is an essential task because different modalities
provide complementary information and joint analysis can provide valuable
insights in clinical practice.

\begin{figure}
	\centering
    \includegraphics[height=2.5cm,width=10cm]{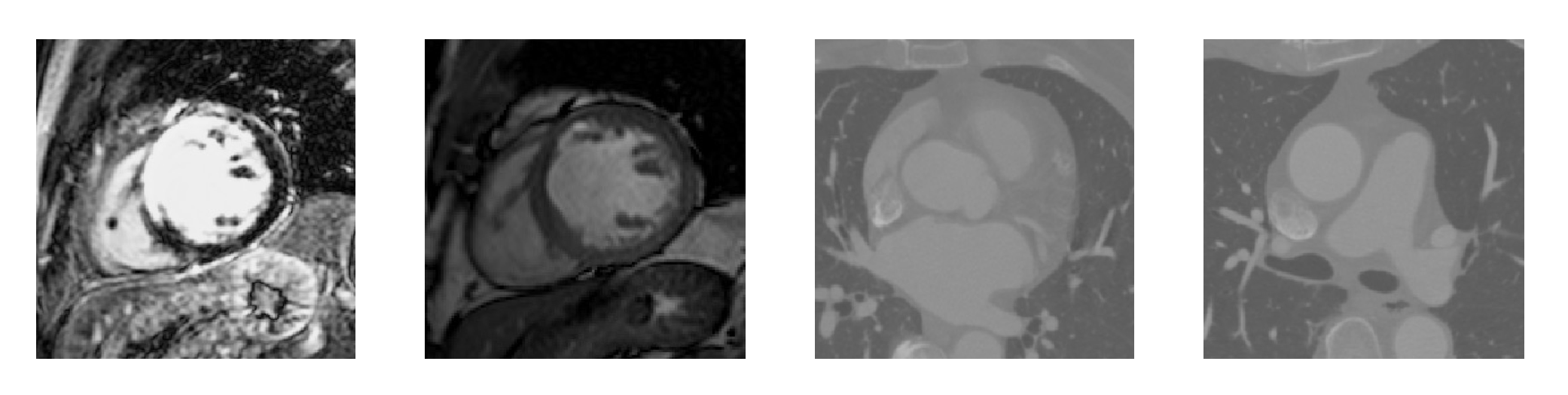}
    \caption{Left to right: slices of MR (left two) and CT (right two) cardiac scans.}
	\label{samples}
\end{figure}

Multi-modal learning is inherently challenging for two reasons:
1) supervised feature learning is often \textit{modality-dependent};
features learned from a single modality can not easily be combined with those from other modalities;
2) joint learning often requires images from different modalities being \textit{spatially} well-aligned and paired; obtaining such training data is itself a costly task and often infeasible.
Fig. \ref{samples} shows sample slices from cardiac scans in different modalities. It can be observed that although they all reveal parts of the heart anatomy, their visual appearances vary.  
Segmentation networks are often sensitive to such
discrepancies, which has become a major obstacle for model generalization across modalities.

Spatial misalignment is another issue. Existing image registration methods are
often infeasible, as the spatial correspondences among modalities can be highly
complex and finding a good similarity measurement is non-trivial.

\begin{figure}[!ht]
	\centering
	\includegraphics[height=2.5cm,width=10cm]{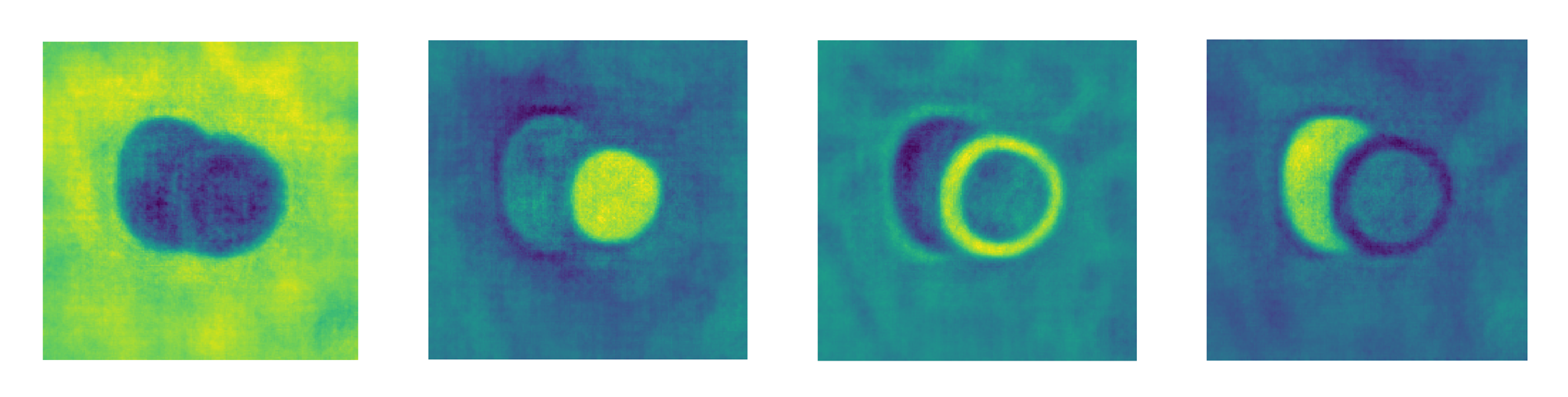}
	\caption{Visualization of feature maps of one cardiac slice in four classes. Left to right: the background, RV, myo and LV. The higher brightness part indicates the region for the corresponding class; it could be observed that class-specific feature representation is brighter than the other parts, which is highly correlated with the ground truth mask in this class. }
	\label{feature}
\end{figure}

To mitigate these issues, joint learning with unpaired data is
emerging as a promising direction
\cite{olterink2017deep,valindria2018multi,dou2020unpaired}.  MultiResUNet
\cite{ibtehaz2020multiresunet} has been proposed to improve upon U-Net in multimodal medical image analysis.  In brain image segmentation, Nie et al.
\cite{nie2016fully} trained networks independently for single modalities and
then fused the high-layer outputs for final segmentation.  Yang et al.
\cite{yang2019unsupervised} used disentangled representations to achieve CT and
MR adaptation.
Existing methods didn't take into account class-specific
information, even though the features obtained by supervised training are highly
correlated with the tasks (Fig. \ref{feature}).

Our assumption is that, with the same network architecture, the underlying
anatomical features should be extracted in a similar manner across modalities.
At the same time, each network instance should have modality-specific modules
to tolerate the imaging domain gaps. With this assumption, to facilitate effective joint feature learning, we adopt an FCN for all modalities, where the
convolutional kernels are shared, while the modality-specific batch feature
normalizations remain local to each modality.  More importantly, we extract
class-specific affinity measurements at multiple feature scales, and minimize
an affinity loss during training.  Different from cross-modal feature consistency loss, our design ensures that the networks extract modality independent features in a similar hierarchical manner. Intuitively, this could
be interpreted as ``high-order'' feature extraction consistency compared with
the feature map consistency loss.  We show that this is a more appropriate
joint model regularizer that effectively guides the anatomical feature
extractions.

In summary, our main contributions are: 1) we propose a novel unpaired multimodal segmentation framework, which explicitly extracts the \textit{modality-agnostic} knowledge; 2) we introduce a joint learning strategy and a \textit{class-specific} affinity matrix to guide the training, which is capable of distilling between-layer relationships at multiple scales; 3) we extensively evaluated our method on two public multimodal benchmark datasets, and the proposed method outperforms the state-of-the-art multimodal segmentation approaches.

\section{Methodology}
This section details the proposed joint training for segmentation tasks. 
\subsection{Multimodal Learning}
\begin{figure}%
	\centering
    \includegraphics[height=4.8cm,width=12cm]{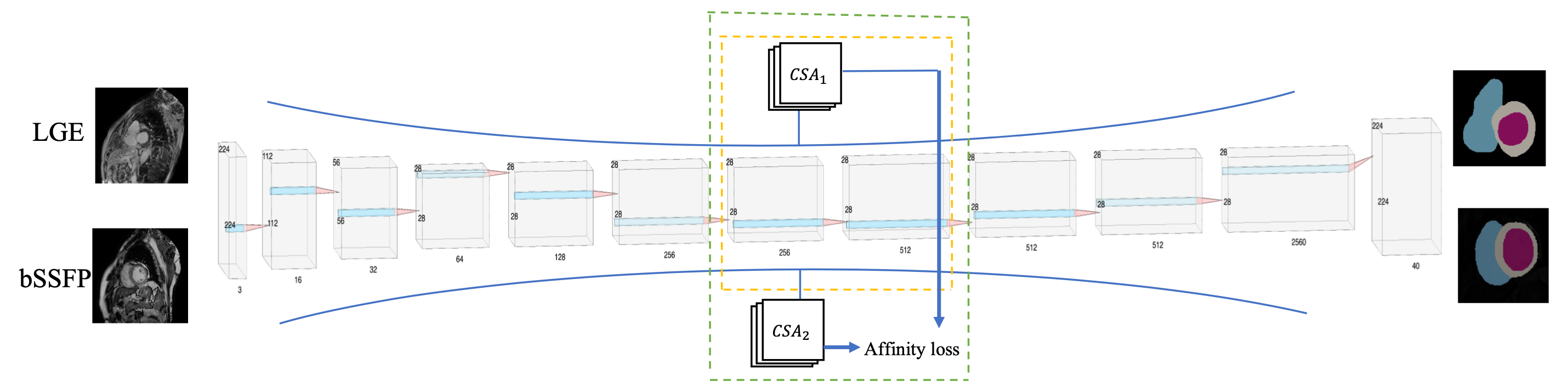}
	\caption{Architecture of the proposed class-specific affinity guided learning for multimodal segmentation (CSA denotes class-specific affinity). The overall structure of the proposed method contains two streams of FCNs with shared layers.  For the detail of the CSA module, please see Fig. \ref{csa_loss}.}
	\label{fig_framework}
\end{figure}

We adopt an FCN as the backbone of our framework. Without any loss of generality, we present our framework in the case of training with two imaging modalities. The overall architecture is illustrated in Fig.~\ref{fig_framework}. The training of the system operates on random unpaired samples from both modalities, the same set of convolutional layers of the network are updated, while the batch normalization layers are initialized and updated individually for each modality.

\subsection{Modality-specific Batch Normalization}
Using two independent sets of parameters for joint training leads to large models and thus tends to overfit.
Karani et al.~\cite{karani2018lifelong} showed that using a domain-specific batch normalization is effective in addressing domain gaps issue while keeping the model compact. Here we employ the same technique for modality-specific feature extraction. Specifically, the batch normalization layer matches the first and the second moment of the distributions of feature x:

\begin{equation}
	x^{*}=\gamma  \frac{x-E(x)}{\sqrt{Var(x)+\epsilon} }+\beta
	\label{equ_bn}
\end{equation}
where $\gamma$, $\beta$ are trainable parameters and are modality-dependent in our design. $\epsilon$ is a very small positive number to avoid dividing by zero. 
\subsection{Class-specific Affinity}

It has been shown that feature maps in a network could reflect the saliency of the class of interest in a multi-class setting \cite{levine2019certifiably} in a recent study by Levine et al. Such a saliency map could give a robust interpretation of the reasoning process of model predictions. Motivated by this study, in the multi-modal segmentation network, since all the modalities share the same tasks (e.g., multi-class heart region segmentation), we hypothesized that the reasoning process of model-specific channels should be similar and its feature map should reflect class-specific saliency (i.e., interpretation of the class of interest). As shown in Fig. \ref{feature}, ideally, the region of interest in a learned feature map should be salient and aligned well with its class-label. Therefore, for a learned feature map $F(l)$ of layer $l$ and a given class $c$, we introduce the \textit{class-specific} feature map $F^c(l)$ defined as   
\begin{equation}
	F^c(l) = F(l) \odot M^c
	\label{equ_csa}
\end{equation}

where $M^c$ denotes the ground truth mask of size $(h,w)$ for class $c$  (reshape to the size of feature map if necessary), and  $\odot$ represents Hadamard product. \\
Suppose that $	F^c_m(l)$  and  $F^c_n(k)$ are the  $m$-th and $n$-th class-specific feature maps from layer $l$ and $k$ respectively, we measure their relationships using an \textit{affinity} defined by their cosine similarity, i.e.,

\begin{equation}
  a^c_{m,n} = \frac{1}{S_c}*cos(F^c_{m}(l),F^c_{n}(k))
   	\label{equ_csa_1}
\end{equation}
Where $S_c$ is the size of the region of interest in $M^c$. Such a normalization is to ensure that the affinity is invariant to the size of the saliency region. 
Suppose that layer $l$ and layer $k$ have $M$ and $N$ number of class-specific feature maps, we construct  the between-layer affinity matrix $A^c_{m,n}$, where the entry at $(m,n)$ is $a^c_{m,n}$. The size of $A^c$ is $M$ by $N$. Since the affinity is computed based class-specific on feature map, we term this as the \textit{class-specific affinity} (CSA) matrix.  

\begin{figure}
	\centering
    \includegraphics[height=3.5cm,width=10cm]{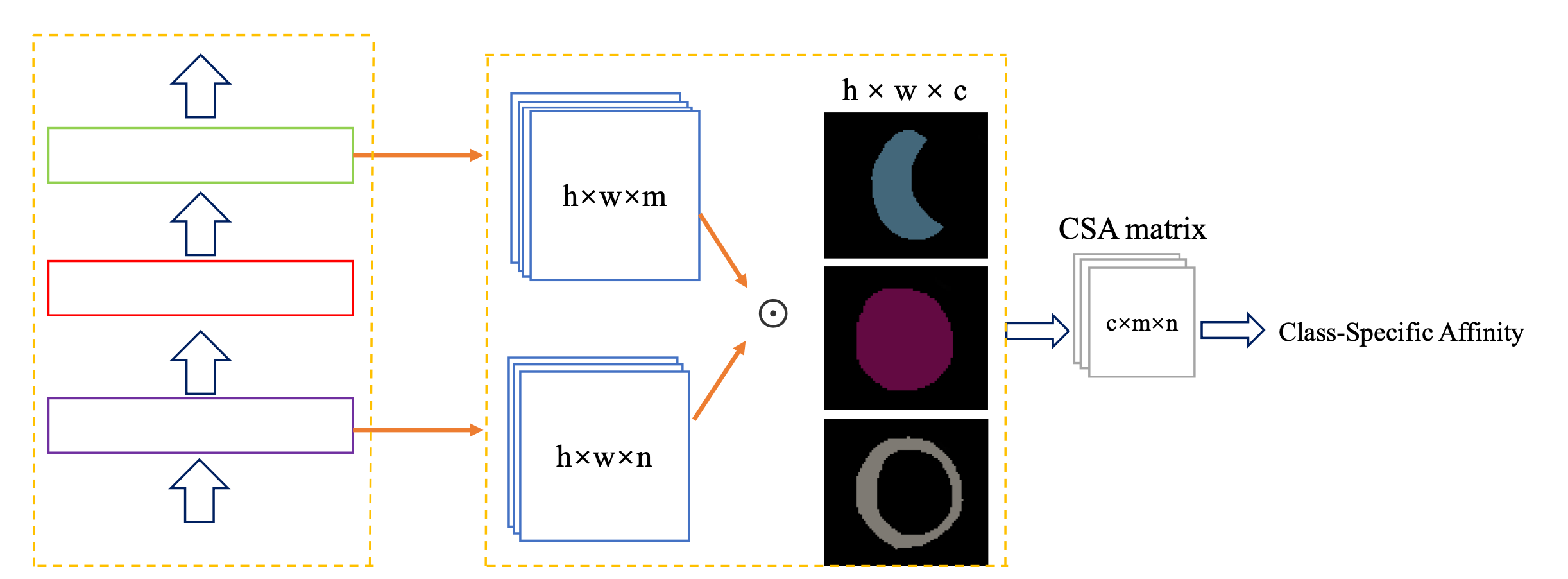}
	\caption{The proposed class-specific affinity guided learning layer. The dotted box on the left shows three convolutional layers for feature extraction. The dotted box in the middle shows the affinity computation by incorporating feature maps at multiple scales as well as the multi-class ground truth segmentation map.}
	\label{csa_matrix}
\end{figure}

Fig.~\ref{csa_matrix} shows our design of a class-specific affinity (CSA) layer. Our CSA could be computed for each of the modality, based on which we build the CSA module.  It is worth noting that our design is based on a class-specific feature map, and independent to the choice of modalities; therefore, our network does not require inputs spatial alignment.

\subsection{CSA Module}

Suppose that we have two modalities using the same network architecture for joint learning. We compute  $A^{c,1}$ for modality-1 (e.g., CT) and $A^{c,2}$ modality-2 (e.g MR) across layer $l$ and $k$. The knowledge encoded by CSA for a specific class $c$ could be transferred by enforcing the consistency of CSA between the two modalities using an L2 norm. We then aggregate all of the consistencies for all of the classes to formulate a consistency loss function as below:   

\begin{figure}[!ht]
	\centering
	\includegraphics[height=3.5cm,width=10cm]{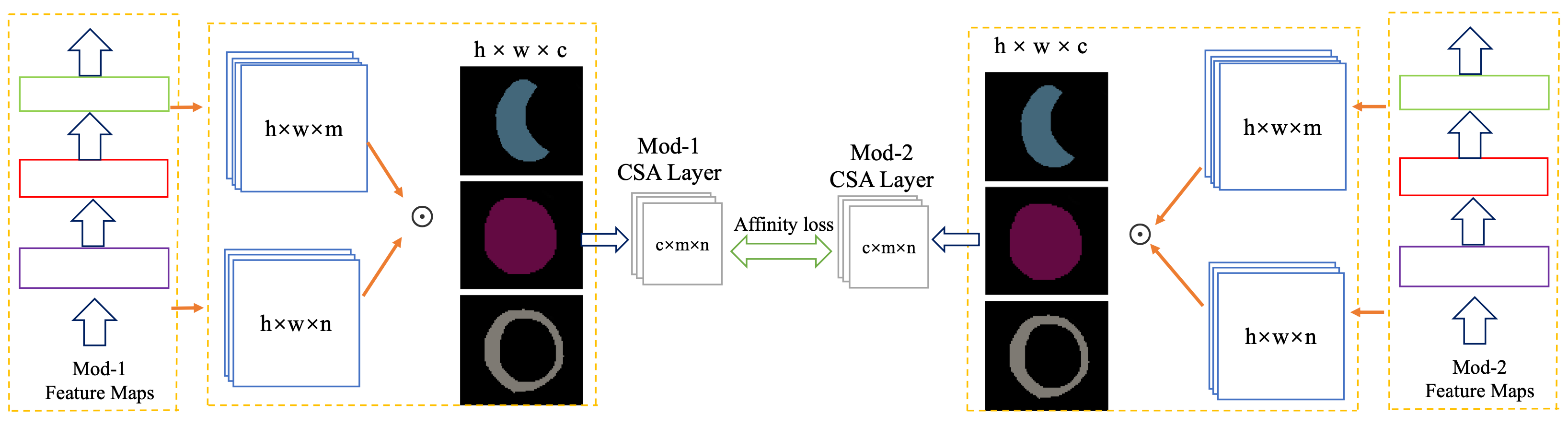}
	\caption{The CSA guided multi-modal knowledge sharing. The CSA Layer computes affinity from "Mod-1" and "Mod-2" respectively (according to Sec. 2.3). The affinity loss between two modalities is computed according to Sec. 2.4.}
	\label{csa_loss}
\end{figure}

\begin{equation}
	L_{CSA}=  \frac{1}{C} \sum_{c}^{C} \left \|\frac{1}{P}\sum_{i=1}^{P}(A_{i}^{c,1} - A_{i}^{c,2})\right \|_{2}^{2}
	\label{equ_relation_loss}
\end{equation}
where $C$ is the number of classes. $P$ is the total number of entries in $A^c$, i.e., $MN$. Normalizing by $P$ is to ensure that the consistency is invariant to the number of feature channels.

When minimizing the CSA loss, the affinity consistency between the two modalities could be maximized thus ensuring joint learning. For the segmentation loss, we use a Dice coefficient loss  $L_{dice}$ to ensure good segmentation at region level, and a $L_{ce}$ cross-entropy loss for pixel-level segmentation.  Taking all of the three losses together, the final loss of our multi-modal learning framework for two modalities 1 and 2 is defined as:

\begin{equation}
	L=\alpha(L_{dice} ^{1} + L_{dice} ^{2}) + \beta (L_{ce} ^{1} +L_{ce} ^{2})  + \lambda L_{CSA}
	\label{equ_loss}
\end{equation}

Where $L_{CSA}$ is the CSA transfer loss, $\alpha$,$\beta$,$\lambda$ are the system parameters to weight the loss components. 

\section{Experiments}

\noindent\textbf{Datasets.} We evaluated our multimodal learning method on two public datasets: MS-CMRSeg 2019 \cite{zhuang2018multcardiac} \cite{zhuang2020cardiac} and MM-WHS \cite{zhuang2019evaluation}. 
MS-CMRSeg 2019 contains 45 patients who had previously suffered from cardiomyopathy. Each patient has images in three cardiac MR modalities (LGE, T2, and bSSFP) with the three segmentation classes: left ventricles, myocardium, and right ventricles. We use LGE and bSSFP in our experiments. For each slice, we crop the image and get region of interests with a fixed bounding box (224$\times$224), enclosing all the annotated regions. we randomly divided the dataset into 80\% for training and 20\% for testing according to the patients.
MM-WHS contains the MR and CT images whole heart from the upper abdomen to the aortic arch. The segmentation task includes four structures: left ventricular myocardium (LVM), left atrium (LAC), left ventricle (LVC), and ascending aorta (AA). We crop each slice with a region with a 256 $\times$ 256-pixel bounding box in the coronal plane and randomly divided the dataset into training (80\%) and test sets (20\%). 
For preprocessing, we use \emph{z-score} normalization \cite{jingkun2020adversarial} to calibrate the intensity of each 2D slice from all modalities in both datasets.

\noindent\textbf{Implementation.} Our architecture consists of nine convolutional operation groups, one deconvolutional group and one softmax layer. Each group contains two to four shared convolutional layers and domain-specific batch normalization layers. 
We implemented the proposed method with Python-based on Tensorflow 1.14.0 library using Nvidia Quadro RTX GPU (24G). We optimize our network with Adam optimizer with a batch size of 8. The learning rate is initialized to 1$\times$ $10^{-4}$ and decayed by 5\% per 1000 iterations. Besides, we incorporated dropout layers (drop rate of 0.75) into the network to validate the performance. In the training stage, we used three loss functions. Empirically, $\alpha$, $\beta$ and $\lambda$ are set to 1, 1 and 0.5 respectively. Our assumption is that the higher layer features are too closely related to the ground truth mask, so the affinity features of the migration process are not obvious, so we choose the feature maps from the intermediate layer. We believe that the information between remote feature maps is difficult to express completely with affinity feature maps, so we used the feature maps from the layers relatively close to each other in our experiment.

\noindent\textbf{Comparison and Analysis.} We designed the following six experimental settings (single training of separate modalities (Single), Unpaired Multi-modal Segmentation via Knowledge Distillation (UMMKD)\cite{dou2020unpaired}, Joint training of two modalities in shared BN model(Joint), Modality-specific batch normalization only (MSBN), Affinity-guided learning (Affinity), and CSA guided learning (ours)). For all settings, the network architecture and datasets are fixed so that different
methods can be compared fairly. In terms of quantitative performance measurements, we adopt the volume Dice score and surface Hausdorff distance as listed in Table \ref{tab:lge_c0}, \ref{tab:dice_ct} and \ref{tab:hausdorff_ct}.

\begin{table}[ht]
\centering
\caption{Average Dice score (\%) and Surface Hausdorff distance(mm) on LGE and bSSFP, the highest performance in each class is highlighted.}
\begin{tabular}{|l|l|l|l|l|l|l|l|l|l|l|l|l|l|l|l|}
\hline
\multirow{2}{*}{Method}      &  \multicolumn{3}{c|}{LGE\_Dice} &\multicolumn{3}{c|}{bSSFP\_Dice} & \multicolumn{3}{c|}{LGE\_Dist.} &\multicolumn{3}{c|}{bSSFP\_Dist.}  \\
\cline{2-13}
 ~ &  LV & myo & RV  & LV & myo & RV &  LV & myo & RV  & LV & myo & RV\\
\hline
Single\cite{long2015fcn} &  90.17 & 80.31 & 86.51  & 92.89 & 85.11 & 88.76 &  6.63 & 3.61 & 25.10 & 47.80 & 2.45 & 11.58\\

UMMKD\cite{dou2020unpaired}  &  90.41 & 80.48 & 86.99  & 93.38 & 85.69 & 89.91 &  5.00 & 3.46 & 8.66  & 3.00 & 2.24 & 5.39\\

Joint &  90.22 & 80.91 & 86.61  & 92.74 & 85.19 & 89.12 &  4.00 & 6.08 & 5.12 & 3.16 & 2.45 & 5.00\\

MSBN &  90.23 & 80.81 & 86.08 & 93.46 & 85.57 & 89.91 &  5.75 & 3.61 & 5.48 & 3.00 & 2.24 & 5.10 \\

Affinity &  90.65 & 81.18 & 87.27  & 93.54 & 85.93 & 89.93 &  5.00 & 3.16 & 9.72 & 3.00 & 2.24 & 5.10\\

Ours &  \textbf{91.89} & \textbf{83.39} & \textbf{87.66}  & 93.48 & 85.72 & \textbf{90.64} & 4.12 & \textbf{3.00} & \textbf{5.00} & \textbf{3.00} & \textbf{2.24} & \textbf{4.12}  \\

\hline
\end{tabular}
\label{tab:lge_c0}
\end{table}

1) Results on MS-CMRSeg 2019: Table \ref{tab:lge_c0} lists the results of three classes cardiac segmentation. As can be observed, compared with individual training and the other multi-modal learning, our methods showed a significant performance gain. The overall \emph{average} Dice score of three classes on two modalities increased to 87.65\% and 89.95\% which is much higher than single training of separate modalities (85.66\% and 88.92\%) and UMMKD (85.96\% and 89.66\%), and the \emph{average} Hausdorff distance of three classes on LGE modality of UMMKD decreased from 5.71mm to 4.04mm. This indicates that class-based cross-modal knowledge transfer is effective. We have conducted the Wilcoxon signed-rank test for the improvements of our method over UMMKD on the results based on the patient-level predictions (the p-value is 0.010 for LGE, and 0.048 for bSSFPP), indicating that improvements are statistically significant. It is observed that the affinity guided learning also achieved the better result on the two modalities, especially on the bSSFP modality, but it is still lower than the CSA module, as the CSA module learned to share the class-based semantic knowledge.

2) Results on MM-WHS: The proposed method was also used to segment six classes cardiac MRI and CT, and achieves promising segmentation performance on this relatively large dataset (Table \ref{tab:dice_ct} and \ref{tab:hausdorff_ct}), with an \emph{average} Dice score of 79.27\% for LVM, 87.46\% for LAC, and 81.42\% for AA on CT modality; 88.89\% for LVM, 92.53\% for LVC, and 96.07\% for AA. Hausdorff distance of our method on the two modalities is also lower in most of the classes. The differences when comparing with the models trained in MS-CMRSeg 2019 were the number of samples for training and the weight of the CSA loss increased to 0.5. Overall, it can be seen that the proposed methods (ours) outperforms both the single modality model and the multi-modal method, which confirms their effectiveness. We tested the CSA learning between the more distant layers and the closer layers, and we found that the performance within the closer layers is better. The CSA knowledge between the closer layers can be better learned and also easier to be migrated.

\begin{table}[!ht]
\centering
\caption{Average Dice score (\%) on MRI and CT.}
\begin{tabular}{|l|l|l|l|l|l|l|l|l|}
\hline
\multirow{2}{*}{Method}  &  \multicolumn{4}{c|}{CT} &\multicolumn{4}{c|}{MRI}   \\
\cline{2-9}
 ~ & LVM & LAC& LVC & AA  & LVM & LAC& LVC & AA \\
\hline
Single\cite{long2015fcn} &  78.17 & 85.84 & 93.07 & 81.08 & 87.56 & 90.29 & 91.66 & 94.26 \\
UMMKD\cite{dou2020unpaired} & 78.73 & 83.47 & 93.29 & 81.41 & 87.89 & 91.68 & 91.88 & 95.32 \\
Joint &  77.92 & 83.96 & 93.51 & 80.08 & 84.03 & 88.41 & 90.92 & 94.67 \\
MSBN  & 78.82 & 86.07 & 94.40 & 80.57 & 87.57 & 92.30 & 91.88 & 96.02 \\
Affinity &  78.63 & 87.13 & 94.35 & 79.49 & 85.31 & 91.06 & 91.84 & 92.24 \\
Ours &  \textbf{79.27} & \textbf{87.46} & 94.22 & \textbf{81.42}  & \textbf{88.89} & 91.18 & \textbf{92.53} & \textbf{96.07} \\
\hline
\end{tabular}
\label{tab:dice_ct}
\end{table}

\begin{table}[!ht]
\centering
\caption{Surface Hausdorff distance(mm) on CT and MRI.}
\begin{tabular}{|l|l|l|l|l|l|l|l|l|}
\hline
\multirow{2}{*}{Method}      &  \multicolumn{4}{c|}{CT} &\multicolumn{4}{c|}{MRI}   \\
\cline{2-9}
 ~ & LVM & LAC& LVC & AA & LVM & LAC& LVC & AA \\
\hline
Single\cite{long2015fcn}&  5.00 & 10.63 & 5.00 & 13.38 & 8.50  & 15.03 & 4.47 & 6.00\\
UMMKD\cite{dou2020unpaired}  & 5.83 & 14.40 & 6.01 & 13.97 & 10.05  & 8.94 & 4.47 & 3.61 \\
Joint &  5.39 & 14.32 & 4.24 & 13.30& 6.40 & 40.17 & 9.84 & 4.47 \\
MSBN  & 5.00 & 10.63 & 4.24 & 13.93 & 4.12 & 9.06 & 4.47 & 3.00 \\
Affinity &  6.00 & 10.63 & 4.89 & 15.13 & 10.05  & 12.08 & 5.00 & 67.09\\
Ours &  5.39 & \textbf{10.30} & \textbf{4.24} & 15.80  & \textbf{4.12} & 9.49 & \textbf{4.12} & \textbf{2.83} \\
\hline
\end{tabular}
\label{tab:hausdorff_ct}
\end{table}

\begin{figure}[!ht]
	\centering
    \includegraphics[height=4cm,width=12cm]{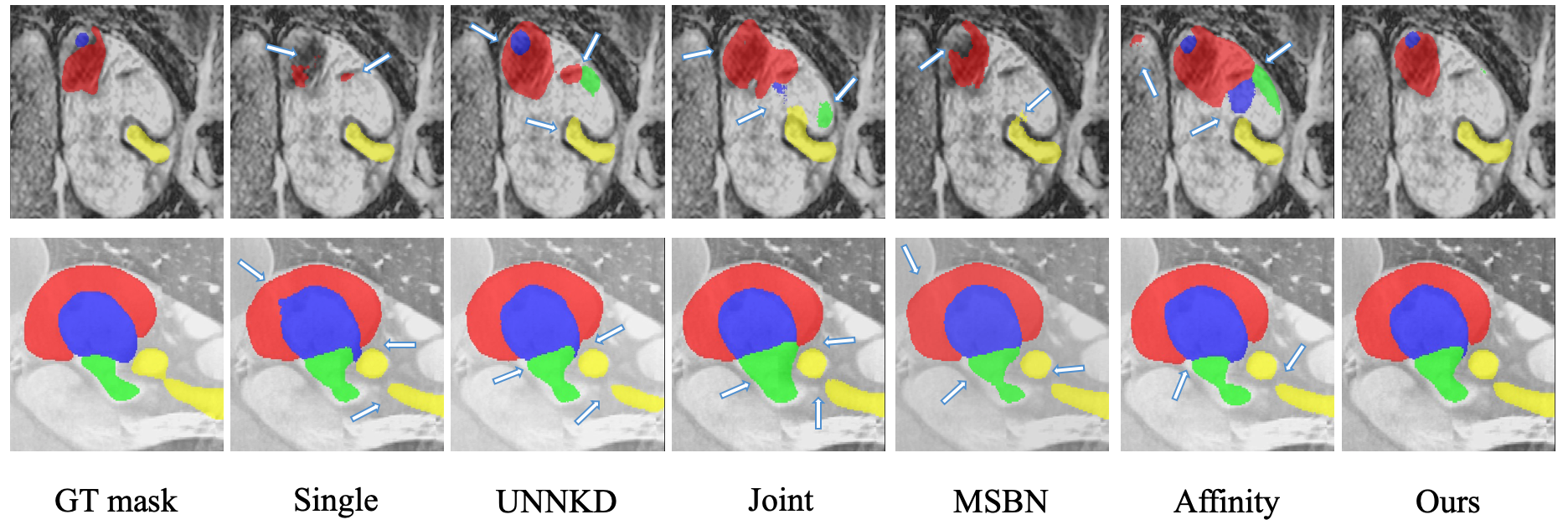}
    \caption{Segmentation maps on CT (top) and MRI (bottom). Left to right: Ground Truth mask, Single, UMMKD, Joint, MSBN, Affinity and CSA.}
	\label{fig:lge_c0}
\end{figure}

\noindent\textbf{Visualization.} Fig. \ref{fig:lge_c0} shows the predicted masks from the six methods. It could be seen that our methods improve the performance of only using single-modal method and the other multi-modal method, especially for the LGE modality. These observations are consistent with those shown in the Table \ref{tab:lge_c0}, \ref{tab:dice_ct} and \ref{tab:hausdorff_ct}.

\section{Conclusion}

We propose a new framework for unpaired multi-modal segmentation, and  introduce class-specific affinity measurements to regularize the jointly model training. We have derived the formulations and experimented with spatially 2D feature maps. As future work, the same concepts could be extended to spatially 3D cases. The results based on the proposed class-specific affinity loss are encouraging.
Further quantitative analysis of the feature maps and investigating model interpretability is also an interesting future direction.

\bibliographystyle{splncs04}
\bibliography{paper2040}

\end{document}